\documentclass[conference]{IEEEtran}

\usepackage{url}
\usepackage{graphicx}
\usepackage{amsmath, amssymb}
\usepackage{booktabs}
\usepackage{hyperref}
\usepackage{bookmark}
\usepackage{subcaption}
\usepackage{xcolor}
\usepackage{algorithm}
\usepackage{algpseudocode}
\usepackage{listings}
\usepackage{cite}
\hypersetup{hidelinks}
\usepackage{multirow}

\lstdefinestyle{py}{
  language=Python,
  basicstyle=\footnotesize\ttfamily,
  numbers=left,
  numberstyle=\tiny,
  stepnumber=1,
  numbersep=6pt,
  frame=single,
  breaklines=true,
  showstringspaces=false,
  keywordstyle=\color{blue},
  commentstyle=\color{gray},
  stringstyle=\color{purple}
}

\hypersetup{
  pdftitle={Optimize, Learn, Refine: Whole-Body Grasping and Pick-and-Throw
with a Spiral Soft Robot},
  pdfauthor={Marwa},
  pdfkeywords={spiral soft robot, actuation-space control, whole-body grasping, CMA-ES, neural warm start, contact-rich manipulation}
}

\IEEEoverridecommandlockouts

\title{\LARGE \bf
Optimize, Learn, Refine: Whole-Body Grasping and Pick-and-Throw
with a Spiral Soft Robot}
\author{}

\author{Marwah Basuhai$^{1}$, Tingcong Liu$^{1,2}$, Ibrahim Alsarraj$^{1}$, Yuhao Wang$^{1}$, Ke Wu$^{1}$ 
}

\begin{document}
\maketitle
\thispagestyle{empty}
\pagestyle{empty}


\begin{abstract}
Soft continuum robots can exploit distributed compliance for whole-body
manipulation, but synthesizing behavior through changing contacts remains
difficult. We address whole-body grasping and pick-and-throw
from an initially ungrasped state through outcome-based actuation-space
optimization. Grasping is quantified by tip angular sweep and body--object
enclosure, while throwing further incorporates release-direction alignment
and minimum release speed. These objectives allow grasping, acceleration,
and release to emerge from compliant interaction without prescribing
contact forces, contact locations, or body configurations. Because the
resulting actuation-to-outcome mapping is nonsmooth, we utilize
derivative-free CMA--ES within an \emph{optimize--learn--refine}
framework. CMA--ES generates solutions for sampled conditions, a
task-conditioned predictor learns warm starts, and CMA--ES refines them
for unseen conditions. In simulation, the method achieves
\(492/500\) successful grasps (\(98.4\%\)) and success rates of
\(98\%\), \(97\%\), and \(94\%\) across three directional throwing
trials. Learned initialization increases grasping success from
\(78.6\%\) to \(98.4\%\) while reducing the median rollout count from
1184 to 816 in CMA--ES. Hardware experiments achieve a \(100\%\)
grasping success rate across 50 executions and a \(100\%\)
pick-and-throw success rate across 30 executions, with 10 repetitions
per direction. Together, these simulation and hardware results
demonstrate the effectiveness of the proposed framework across both
simulated and physical whole-body manipulation tasks.
\end{abstract}

\section{Introduction}
Soft continuum robots can conform to object geometry and distribute
contact over extended portions of their deformable bodies
\cite{mehrkish2021taxonomy}.
Their compliance also allows local interaction variations to be
accommodated passively \cite{hauser2023morphological}.
These capabilities have motivated interest in whole-body manipulation,
in which coordinated body deformation and distributed contact are
used to acquire and manipulate objects.

Whole-body grasping is a fundamental and widely studied form of
soft whole-body manipulation. Existing methods determine feasible
wrapping configurations \cite{li2011graspconfig}, plan
surface-following motions \cite{li2016progressive}, or synthesize
robot configurations and contact solutions
\cite{mehrkish2022synthesis,graule2022contactimplicit}.
Complementary control methods stabilize desired whole-body grasps
\cite{chu2023fullbody}. More dynamic manipulation, however, requires going beyond
object acquisition and retention. In pick-and-throw,
the grasp should be maintained during acceleration and then released
to produce the desired object motion.
Existing throwing systems provide object retention in different ways.
TossingBot combines parallel-jaw grasping with learned throwing
primitives \cite{zeng2020tossingbot}, while soft-arm studies use
a kinematic end-effector attachment \cite{bianchi2022open},
a distal or vacuum gripper
\cite{bianchi2024softoss,bianchi2024softsling},
or a prescribed initial body grasp
\cite{morimoto2022characterization}.

Across these formulations, object retention is established through
a separate gripper, a kinematic end-effector attachment, or a prescribed
initial body grasp. The
throwing phase therefore begins from a secured object, rather than including whole-body grasp acquisition, contact-mediated acceleration, and release from an initially ungrasped state within the same interaction.

\begin{figure*}[!t]
    \centering
    \includegraphics[width=0.96\textwidth]{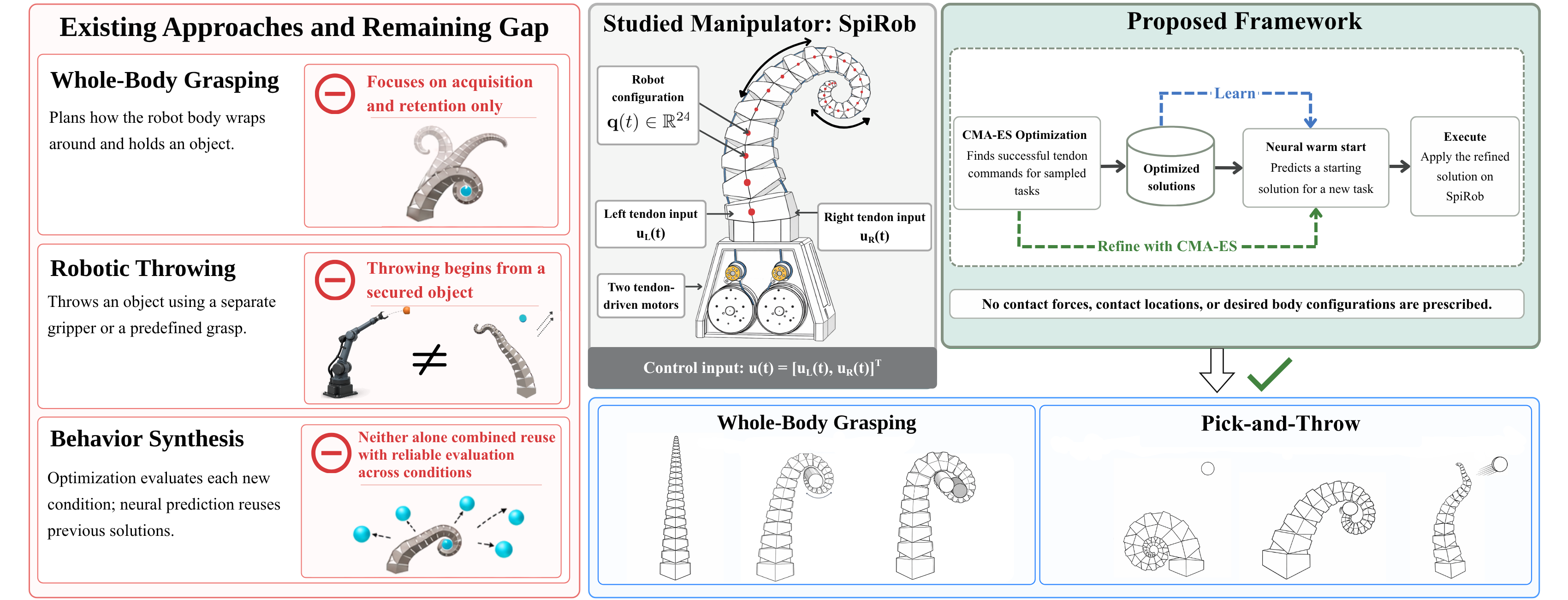}
    \caption{Overview of the motivation  and proposed
    optimize--learn--refine framework for soft robotic whole-body manipulation.}
    \label{fig:intro_framework}
\end{figure*}

This setting makes control particularly difficult.
As the interaction progresses from acquisition and retention to
acceleration and release, changing contact modes couple tendon
actuation with body deformation, friction, and object motion
\cite{dellasantina2023softcontrol}, resulting in a potentially
nonsmooth actuation-to-outcome mapping. Differentiable simulation has enabled trajectory optimization under
structured contact models
\cite{bern2019trajectory,menager2025differentiable}, while black-box
search avoids explicit contact derivatives by evaluating complete
interaction episodes \cite{zwane2024dynamic}.
However, independently optimizing each new task condition requires
repeated search and does not exploit the relationship between new
conditions and previously successful solutions.
Learning-based methods can instead capture such cross-condition
regularities \cite{shentu2026rigidbody}, but directly predicted commands
may not reproduce the desired outcome when the contact interaction
changes under a new condition. These limitations make it difficult to efficiently generate
reliable contact-rich behaviors across varying task conditions.
To address these limitations summarized in Fig.~\ref{fig:intro_framework}, we develop an outcome-based
optimize--learn--refine framework for whole-body grasping and pick-and-throw with soft robots.
The framework couples task-level outcome specification with
cross-condition behavior reuse and condition-specific actuation search.
The main contributions of this work are threefold:

\begin{itemize}

\item \textbf{A shared outcome-based formulation for whole-body enclosure and directional release.} Grasping is formulated as terminal whole-body enclosure and extended to pick-and-throw via release-direction and minimum-speed requirements.

\item \textbf{An optimize--learn--refine strategy for contact-rich behavior synthesis.} CMA--ES optimizes tendon commands through rollouts, while a neural model reuses prior solutions to initialize refinement for new conditions.

\item \textbf{Simulation and hardware validation.} In simulation, the method achieves a 98.4\% grasping success rate and
pick-and-throw success rates of 98\%, 97\%, and 94\% across three
throwing directions. On hardware, it achieves 100\% success for both
whole-body grasping and pick-and-throw, with 50 grasping executions and
30 throwing executions comprising 10 repetitions per direction.
\end{itemize}

\section{Problem Statement}

This section defines the grasping and pick-and-throw tasks considered in
this work and introduces the tendon-driven soft robot SpiRob.

\subsection{The Studied Tendon-Driven Soft Robot: SpiRob}
\label{studied-manipulator}

We use SpiRob, the spiral-inspired tendon-driven soft robot shown in the
central panel of Fig.~\ref{fig:intro_framework}, as the experimental platform
\cite{wang2024spirobs}. Fabricated from flexible polymers through additive
manufacturing, SpiRob has a logarithmic-spiral-inspired backbone whose
curvature and stiffness vary continuously from a relatively rigid base to a
highly compliant tip. This graded structure provides a large reachable
workspace and supports reaching, wrapping, and grasping motions.

\subsection{Manipulation Task Objectives}

We consider two planar contact-rich tasks starting from a specified,
initially ungrasped object position. In whole-body grasping, SpiRob
approaches the object and forms a compact terminal enclosure.
Pick-and-throw extends this behavior by retaining and
accelerating the enclosed object before releasing it along a prescribed
direction with at least a specified speed. For either task, a
finite-horizon tendon-command trajectory is executed open loop at the task level, without object-state or
contact-feedback updates.

\subsection{Problem Formulation}
We seek a finite-horizon tendon-command trajectory that produces the prescribed manipulation outcome. The two tendon inputs are collected as
\begin{equation}
\label{eq:control_input}
\mathbf{u}(t)
=
\begin{bmatrix}
u_L(t)\\
u_R(t)
\end{bmatrix},
\quad t\in[0,T],
\end{equation}
where $u_L(t)$ and $u_R(t)$ are the left and right tendon inputs,
respectively, and $T$ is the execution horizon.
For a given command trajectory, the robot and object motions are
determined by a complete physics-simulation rollout. We therefore
formulate each task as
\begin{equation}
\label{eq:functional_opt}
\begin{aligned}
\mathbf{u}^{\star}(\cdot)
&\in
\operatorname*{arg\,min}_{\mathbf{u}(\cdot)}
J\bigl(\mathbf{q}(\cdot),\mathbf{P}_o(\cdot)\bigr)
\\
&\textnormal{s.t.}\quad
\mathcal{S}\bigl(
\mathbf{q}(\cdot),
\mathbf{P}_o(\cdot),
\mathbf{u}(\cdot)
\bigr)
=
\mathbf{0},
\end{aligned}
\end{equation}
where $\mathbf{q}(t)\in\mathbb{R}^{n}$ is the robot configuration,
$\mathbf{P}_o(t)\in\mathbb{R}^{2}$ is the object-center position,
$J$ is the task-dependent outcome cost, and
$\mathcal{S}$ denotes the robot--object--environment simulation
equations over $[0,T]$.
Thus, the tendon-command trajectory is optimized directly, while
the robot motion, object motion, and contact evolution arise from
the simulated interaction.


\begin{figure*}[t]
\centering
\includegraphics[width=0.9\textwidth]{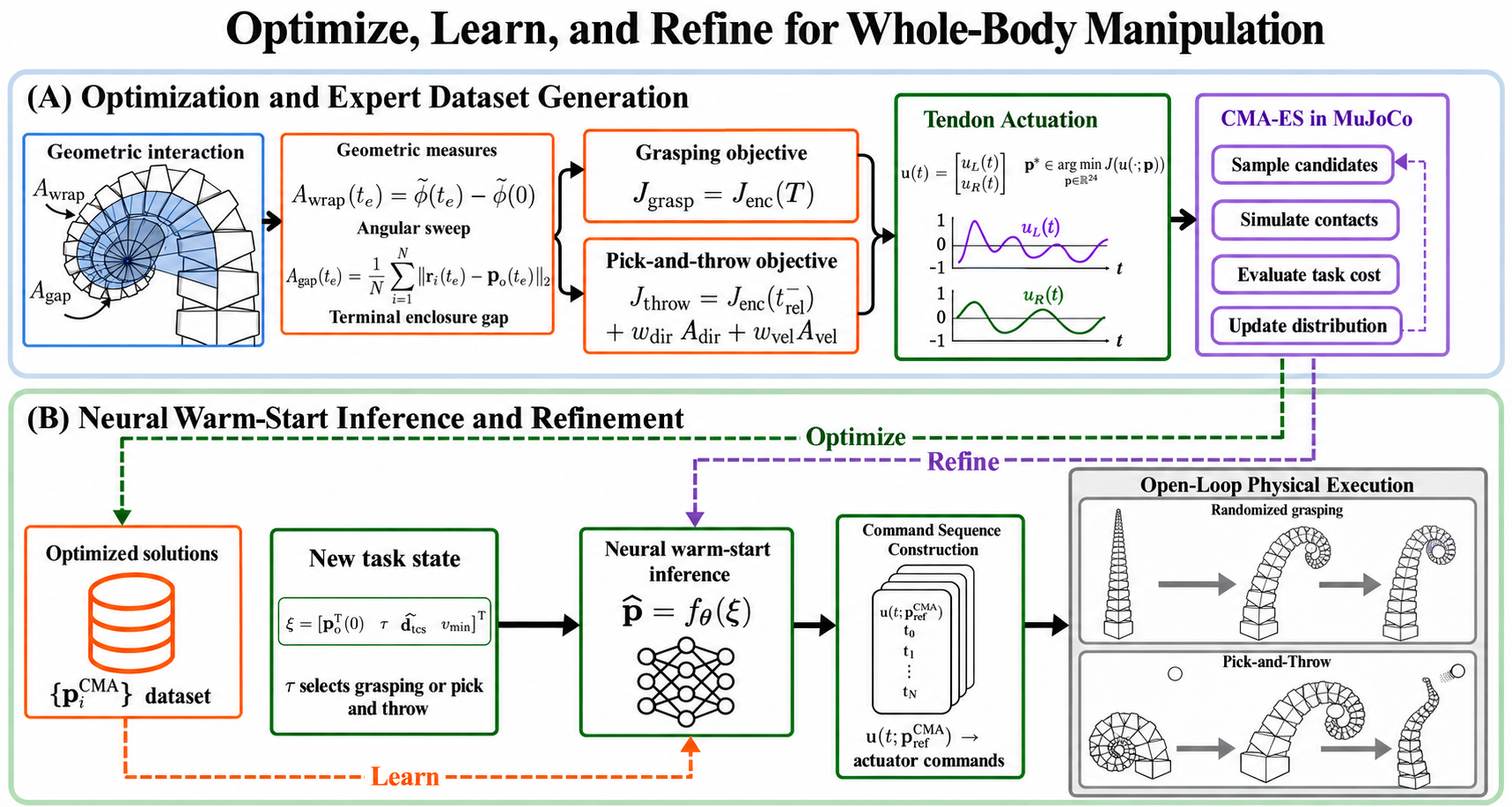}
\caption{Methodological pipeline from task objectives to physical execution.}
\label{fig:methodology_pipeline}
\end{figure*}

\section{Methodology}
\label{sec:methodology}

As shown in Fig.~\ref{fig:methodology_pipeline}, the proposed methodology specifies the three components needed to
solve the actuation-space problem in Eq.~\eqref{eq:functional_opt}.
First, we define the task-dependent outcome cost \(J\) for whole-body
grasping and pick-and-throw.
Second, we represent the continuous tendon-command trajectory
\(\mathbf{u}(\cdot)\) with a finite-dimensional parameter vector for
numerical search.
Finally, we solve the parameterized problem through an
optimize--learn--refine strategy that reuses prior solutions across task conditions. The overall methodological pipeline is shown in Fig.~\ref{fig:methodology_pipeline}.

\subsection{Outcome Costs and Actuation Parameterization}

Whole-body grasping and pick-and-throw share an
enclosure component in the task cost $J$. For grasping, enclosure
is evaluated at task completion, whereas for pick-and-throw it is
evaluated immediately before release together with the released
object's motion.

\paragraph{Whole-body grasping cost}

A whole-body enclosure requires the robot not only to progress
around the object, but also to keep its body close to the object.
We therefore use two complementary geometric proxies.
Tip angular sweep captures progression around the object, whereas
body--object proximity discourages solutions in which the tip moves
around the object while the body remains far away. Let $\mathbf r_{\mathrm{tip}}(t)\in\mathbb R^2$ and
$\mathbf P_o(t)\in\mathbb R^2$ denote the robot-tip and moving
object-center positions in the $x$--$z$ plane, respectively.
To quantify around-object progression, we first define the polar
angle of the tip about the object center as
\begin{equation}
\label{eq:phi}
\phi_{\mathrm{raw}}(t)
=
\operatorname{atan2}
\left(
r_{\mathrm{tip},z}(t)-P_{o,z}(t),
r_{\mathrm{tip},x}(t)-P_{o,x}(t)
\right).
\end{equation}
Because $\phi_{\mathrm{raw}}(t)\in(-\pi,\pi]$, directly differencing
the sampled angles can introduce artificial $2\pi$ jumps.
We therefore unwrap the angle trajectory to obtain the continuous
angle $\widetilde{\phi}(t)$.
For an evaluation time $t_e\in[0,T]$, the resulting signed angular
sweep is
\begin{equation}
\label{eq:Awrap}
A_{\mathrm{wrap}}(t_e)
=
\widetilde{\phi}(t_e)-\widetilde{\phi}(0).
\end{equation}
A larger positive value indicates greater progression of the tip
around the object. Angular progression alone does not ensure that the robot body
remains close to the object.
We therefore measure body--object proximity using sampled points
along the robot body.
Let $\mathbf r_i(t)\in\mathbb R^2$, $i=1,\ldots,N$, denote the
position of the $i$th body sample.
The mean body-to-object-center distance is
\begin{equation}
\label{eq:Agap}
A_{\mathrm{gap}}(t_e)
=
\frac{1}{N}
\sum_{i=1}^{N}
\left\|
\mathbf r_i(t_e)-\mathbf P_o(t_e)
\right\|_2 .
\end{equation}
Smaller values indicate that the robot body remains closer to
the object. We combine these two quantities into the enclosure cost
\begin{equation}
\label{eq:Jenc}
J_{\mathrm{enc}}(t_e)
=
-w_{\mathrm{wrap}}A_{\mathrm{wrap}}(t_e)
+w_{\mathrm{gap}}A_{\mathrm{gap}}(t_e),
\end{equation}
where $w_{\mathrm{wrap}},w_{\mathrm{gap}}>0$ balance around-object
progression and body--object proximity.
Minimizing $J_{\mathrm{enc}}$ therefore favors configurations that
both progress around the object and remain close to it. For whole-body grasping, enclosure is evaluated at the end of the
actuation horizon:
\begin{equation}
\label{eq:Jgrasp}
J_{\mathrm{grasp}}
=
J_{\mathrm{enc}}(T).
\end{equation}

\begin{figure}[!]
\centering
\includegraphics[width=\columnwidth]{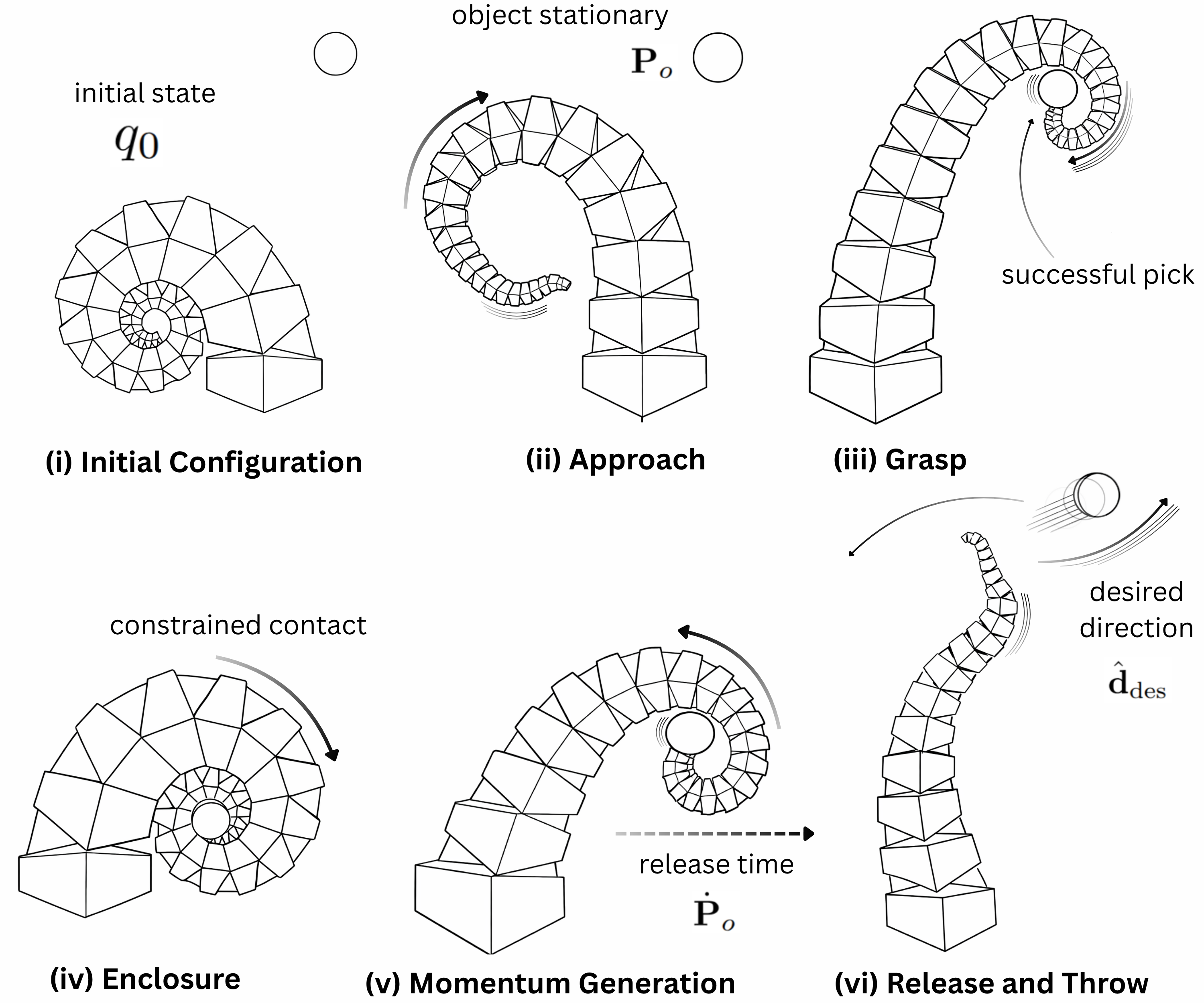}
\caption{Pick-and-throw sequence comprising initial configuration, approach, grasp, enclosure, momentum generation, and release along the desired direction.}
\label{fig:throw_schematic}
\end{figure}

\paragraph{Pick-and-throw cost}

Pick-and-throw retains the enclosure objective in
Eq.~\eqref{eq:Jenc}, but evaluates it immediately before release,
while the post-release object motion determines throwing performance. We define the release time $t_{\mathrm{rel}}$ as the final
robot--object contact-to-separation transition in the simulated
rollout (Fig.~\ref{fig:throw_schematic}). The states immediately
before and after release are denoted by $t_{\mathrm{rel}}^{-}$ and
$t_{\mathrm{rel}}^{+}$, corresponding to the last in-contact and
first subsequent contact-free states, respectively. Rollouts without
such a transition are assigned a fixed failure penalty.

The release outcome is characterized by the object velocity
immediately after separation:
\begin{equation}
\label{eq:vrel}
\mathbf v_{\mathrm{rel}}
=
\dot{\mathbf P}_o(t_{\mathrm{rel}}^{+}).
\end{equation}
A directional throw requires this velocity to be aligned with the desired
throwing direction while having sufficient magnitude. We therefore define
separate direction and speed terms:
\begin{align}
A_{\mathrm{dir}}
=
1-
\frac{
\mathbf v_{\mathrm{rel}}^{\mathsf T}
\hat{\mathbf d}_{\mathrm{des}}
}{
\|\mathbf v_{\mathrm{rel}}\|_2+\delta
},\ 
A_{\mathrm{vel}}
=
v_{\min}-\|\mathbf v_{\mathrm{rel}}\|_2,
\label{eq:Avel}
\end{align}
where $\hat{\mathbf d}_{\mathrm{des}}\in\mathbb R^2$ is the desired
unit release direction, $v_{\min}$ is the prescribed release-speed
reference, and $\delta>0$ prevents division by zero.
The direction term decreases as the release velocity aligns with
$\hat{\mathbf d}_{\mathrm{des}}$, while the speed term decreases as the
release speed increases. The complete pick-and-throw objective therefore combines the shared
enclosure criterion immediately before release with the direction and
magnitude of the release velocity:
\begin{equation}
\label{eq:Jthrow}
J_{\mathrm{throw}}
=
J_{\mathrm{enc}}(t_{\mathrm{rel}}^{-})
+w_{\mathrm{dir}}A_{\mathrm{dir}}
+w_{\mathrm{vel}}A_{\mathrm{vel}}.
\end{equation}
Here, $w_{\mathrm{dir}},w_{\mathrm{vel}}>0$ balance the release-direction
and speed terms, and all objective weights remain fixed across experiments.

\paragraph{Actuation parameterization}

Equation~\eqref{eq:functional_opt} treats the tendon-command
trajectory $\mathbf u(\cdot)$ as the optimization variable.
For finite-dimensional rollout-based search, we represent this
trajectory by a compact and smooth parameter vector $\mathbf p$
that captures coordinated shared and differential variation of
the two tendon commands:
\begin{equation}
\label{eq:pvector}
\mathbf{p}
=
\left[
b_0,
\boldsymbol{\beta}^{\mathsf{T}},
s_0,
\boldsymbol{\gamma}^{\mathsf{T}},
\kappa,
\epsilon,
k_w,
c_w,
\boldsymbol{\rho}^{\mathsf{T}},
\mathbf{a}^{\mathsf{T}}
\right]^{\mathsf{T}}
\in\mathbb{R}^{24}.
\end{equation}
Rather than parameterizing the two tendon commands independently, we decompose them into shared and differential components. Here, $b_0$ and $s_0$ are the corresponding signal offsets, while $\boldsymbol{\beta},\boldsymbol{\gamma}\in\mathbb R^4$
contain their fourth-order polynomial coefficients. The scalar $\kappa$ scales the differential component,
$\epsilon$ introduces a constant differential bias, and $k_w$ and $c_w$ control time warping. The vectors $\boldsymbol{\rho},\mathbf a\in\mathbb R^5$ parameterize the smooth modulation function $g(t)$, constructed as a weighted combination of five smooth phase functions, where $\boldsymbol{\rho}$ determines the phase-transition structure and $\mathbf a$ determines the signed phase amplitudes.

Let
\begin{equation}
\eta(t)=\zeta(t)-\frac{1}{2},
\end{equation}
Here, $\zeta(t)$ is a normalized sigmoid time warp obtained by mapping
$\sigma(k_w(t/T-c_w))$ to $[0,1]$, where
$\sigma(x)=(1+e^{-x})^{-1}$; $k_w$ controls the warp strength and
$c_w$ its center. Thus, $\eta(t)=\zeta(t)-1/2\in[-1/2,1/2]$.
The shared and differential signals are
\begin{align}
b(t)
&=
b_0+
\sum_{k=1}^{4}
\frac{\beta_k}{k!}\eta(t)^k,
\qquad
s(t)
=
s_0+
\sum_{k=1}^{4}
\frac{\gamma_k}{k!}\eta(t)^k.
\end{align}
The shared component $b(t)$ changes both tendon commands together,
whereas the differential component controls their relative actuation.
The latter is further shaped by the smooth modulation function $g(t)$. The resulting left- and right-tendon commands are
\begin{equation}
\label{eq:tendon_commands}
\begin{aligned}
u_L(t;\mathbf{p})
&=
\operatorname{clip}_{[0,1]}
\left(
b(t)+\kappa g(t)s(t)+\epsilon
\right),\\
u_R(t;\mathbf{p})
&=
\operatorname{clip}_{[0,1]}
\left(
b(t)-\kappa g(t)s(t)-\epsilon
\right).
\end{aligned}
\end{equation}
Thus, before clipping, $b(t)$ contributes equally to both tendon
commands, while $\kappa g(t)s(t)+\epsilon$ contributes with opposite
signs. The clipping operation enforces the admissible command range. This parameterization converts Eq.~\eqref{eq:functional_opt} into
the finite-dimensional search
\begin{equation}
\label{eq:param_opt}
\mathbf{p}^{\star}
\in
\operatorname*{arg\,min}_{\mathbf{p}\in\mathbb{R}^{24}}
\mathcal{J}(\mathbf{p}),
\end{equation}
where $\mathcal{J}(\mathbf p)$ denotes $J_{\mathrm{grasp}}$ or
$J_{\mathrm{throw}}$ evaluated on the complete simulation rollout
induced by $\mathbf u(\cdot;\mathbf p)$. The optimization and expert-data generation stage is illustrated in Fig.~\ref{fig:methodology_pipeline}(A).

\subsection{Optimize--Learn--Refine Behavior Synthesis}

The parameterization above converts the original functional problem
into a finite-dimensional search over $\mathbf p$. Since the desired
behavior varies with the initial object configuration and task
requirements, we represent each manipulation condition as
\begin{equation}
\label{eq:task_condition}
\boldsymbol{\xi}
=
\begin{bmatrix}
\mathbf{P}_o^{\mathsf T}(0) &
\tau &
\hat{\mathbf d}_{\mathrm{des}}^{\mathsf T} &
v_{\min}
\end{bmatrix}^{\mathsf T},
\end{equation}
where $\tau\in\{0,1\}$ identifies whole-body grasping or pick-and-throw. The throwing-specific entries
$\hat{\mathbf d}_{\mathrm{des}}$ and $v_{\min}$ are set to zero for
grasping. For each training condition $\boldsymbol{\xi}_i$, we first optimize
the actuation parameters using CMA--ES \cite{hansen2016cmaes}.
The rollout cost is potentially nonsmooth because it depends on
contact-rich robot--object interaction, making derivative-free search
suitable for this problem. Candidate parameter vectors are evaluated
through complete simulation rollouts, and the best-found solution
within the search budget is retained as
$\mathbf p_i^{\mathrm{CMA}}$. Repeating this process over $M$ task
conditions produces
\begin{equation}
\label{eq:optimized_dataset}
\mathcal{D}
=
\left\{
\left(
\boldsymbol{\xi}_i,
\mathbf p_i^{\mathrm{CMA}}
\right)
\right\}_{i=1}^{M}.
\end{equation}
The optimized solutions in $\mathcal D$ capture how effective
actuation varies across task conditions. The neural warm-start inference and condition-specific refinement stage is shown in Fig.~\ref{fig:methodology_pipeline}(B). Rather than restarting the
search from scratch for every new condition, we train a neural
predictor by supervised regression to estimate an actuation-parameter
initialization:
\begin{equation}
\label{eq:nn_predict}
\widehat{\mathbf p}
=
f_{\boldsymbol{\theta}}(\boldsymbol{\xi}),
\end{equation}
where $\boldsymbol{\theta}$ denotes the network parameters. Because $\widehat{\mathbf p}$ is a learned estimate rather than a
solution evaluated under the new contact interaction, it is used to
initialize a condition-specific CMA--ES search instead of being
executed directly. For a new task condition, we set
\begin{equation}
\label{eq:nn_init}
\mathbf m_0
=
\widehat{\mathbf p},
\quad
\mathbf C_0
=
\mathbf I,
\end{equation}
and reevaluate candidate parameters through simulation rollouts under
that condition. CMA--ES then refines the initialization according to
the corresponding task cost and returns the best-found parameter
vector $\mathbf p_{\mathrm{ref}}^{\mathrm{CMA}}$. The resulting
tendon-command trajectory
$\mathbf u(t;\mathbf p_{\mathrm{ref}}^{\mathrm{CMA}})$ is finally used
for task-level open-loop execution.

\section{Optimization and Learning in Simulation}
\label{sec:simulation_validation}




\subsection{Simulation Setup and Data Collection}
\label{subsec:simulation_protocol}

The framework was evaluated in MuJoCo for randomized whole-body
grasping and pick-and-throw. The grasping task used
Eq.~\eqref{eq:Jgrasp}, whereas pick-and-throw used
Eq.~\eqref{eq:Jthrow}. All simulation rollouts used a
$3.5~\mathrm{s}$ control horizon. 

For whole-body grasping, the 84 training positions and 500 test
positions were sampled from the same robot-centered workspace shown in Fig.~\ref{fig:workspace}, spanning
\[
P_{o,x}(0)\in[-0.228,0.227]~\mathrm{m},
\quad
P_{o,z}(0)\in[-0.100,0.450]~\mathrm{m}.
\]
For each training position, CMA--ES generated an optimized
condition--parameter pair
$(\boldsymbol{\xi}_i,\mathbf p_i^{\mathrm{CMA}})$ for neural training.
The manipulated object was modeled as a cylinder of radius
$R=15~\mathrm{mm}$.

\begin{table}[!]
\centering
\caption{Simulation training and test data.}
\label{tab:sim_data_split}
\small
\begin{tabular}{@{}lcc@{}}
\toprule
\textbf{Task} & \textbf{Training} & \textbf{Testing} \\
\midrule
Whole-body grasping
& 84 positions
& 500 new positions \\
Pick-and-throw
& 30 conditions
& 300 evaluations \\
\bottomrule
\end{tabular}
\end{table}


For pick-and-throw, three task settings were considered,
with prescribed release directions of $135^\circ$, $45^\circ$, and
$90^\circ$, respectively. Each setting contributed 10
CMA--ES-optimized training conditions and was evaluated through
100 simulation evaluations. The training and test sets for the two tasks are summarized in
Table~\ref{tab:sim_data_split}.

\subsection{Evaluation Protocol and Compared Methods}
\label{subsec:evaluation_protocol}

For randomized grasping, a trial was considered successful when the robot reached the target and maintained whole-body enclosure at the end of the control horizon. For pick-and-throw, success required a release-direction error below $5^\circ$ and a release speed no lower than the prescribed $v_{\min}$, set to $0.18$, $0.32$, and $0.24~\mathrm{m\,s^{-1}}$ for Trials 1--3, respectively. We compare the proposed pipeline with four alternative optimizer--NN combinations. For a fair comparison, all methods were evaluated under the same task settings and test conditions, using the same neural warm start and optimization budget.

\begin{itemize}
    \item \textbf{CMA--ES\cite{hansen2001cmaes} (Ours):}
    CMA--ES refines the neural initialization through
    population-based covariance adaptation.

    \item \textbf{DE\cite{storn1997differential}:}
    Differential Evolution (DE) refines the neural initialization
    through population-based differential mutation and recombination.

    \item \textbf{SPSA\cite{spall1992spsa}:}
    Simultaneous Perturbation Stochastic Approximation (SPSA)
    refines the neural initialization using perturbation-based
    estimates of the local search direction.

    \item \textbf{TuRBO-1\cite{eriksson2019turbo}:}
    TuRBO-1 refines the neural initialization through local
    Bayesian optimization within an adaptive trust region.

    \item \textbf{Nelder--Mead\cite{nelder1965simplex}:}
    Nelder--Mead refines the neural initialization through
    simplex-based derivative-free search.
\end{itemize}

\begin{figure}[t]
\centering

\begin{subfigure}[t]{\columnwidth}
    \centering
    \includegraphics[width=\columnwidth]{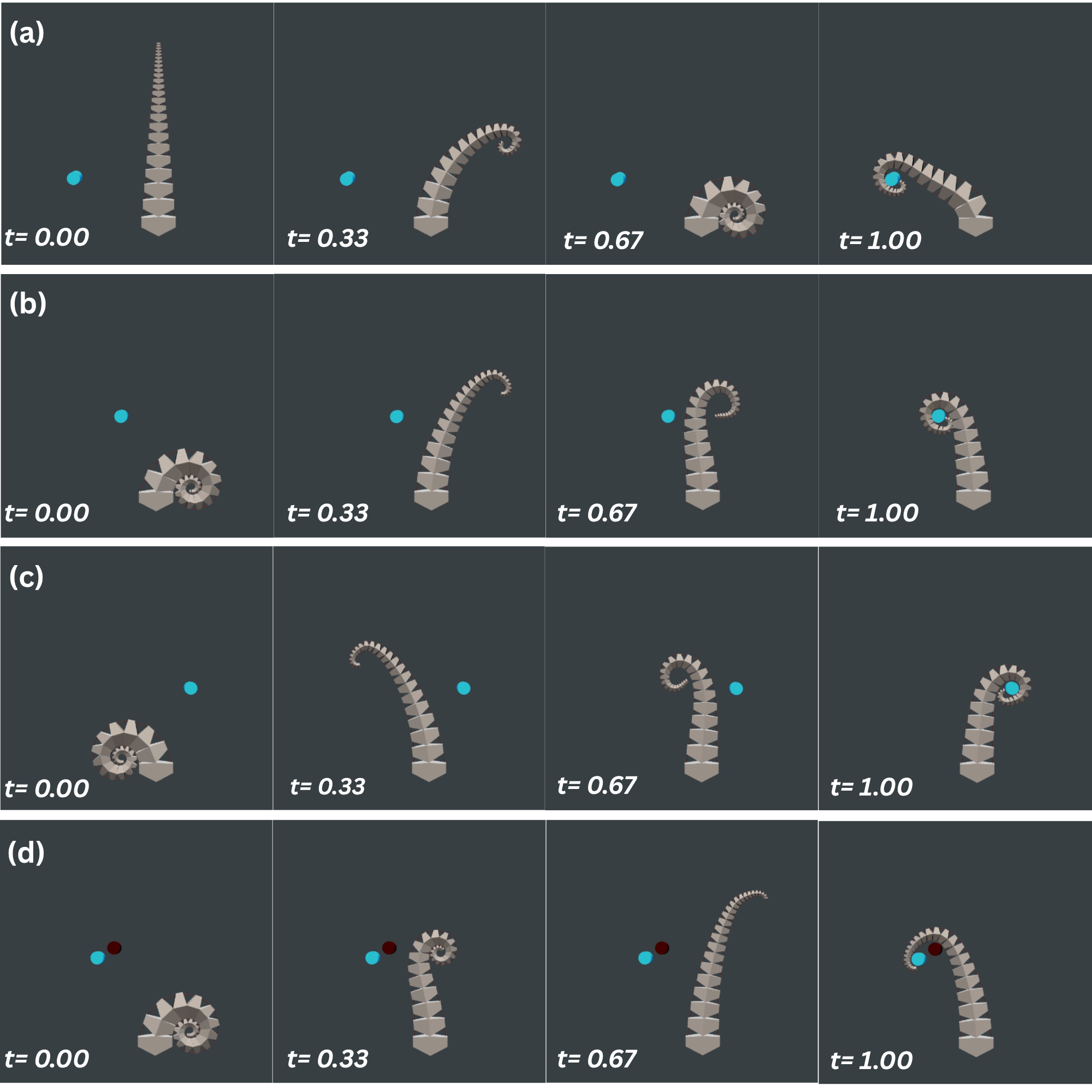}
    \caption{Simulated whole-body grasping trials.}
    \label{fig:simulation_grasping_sequences}
\end{subfigure}

\vspace{2mm}

\begin{subfigure}[t]{\columnwidth}
    \centering
    \includegraphics[width=\columnwidth]{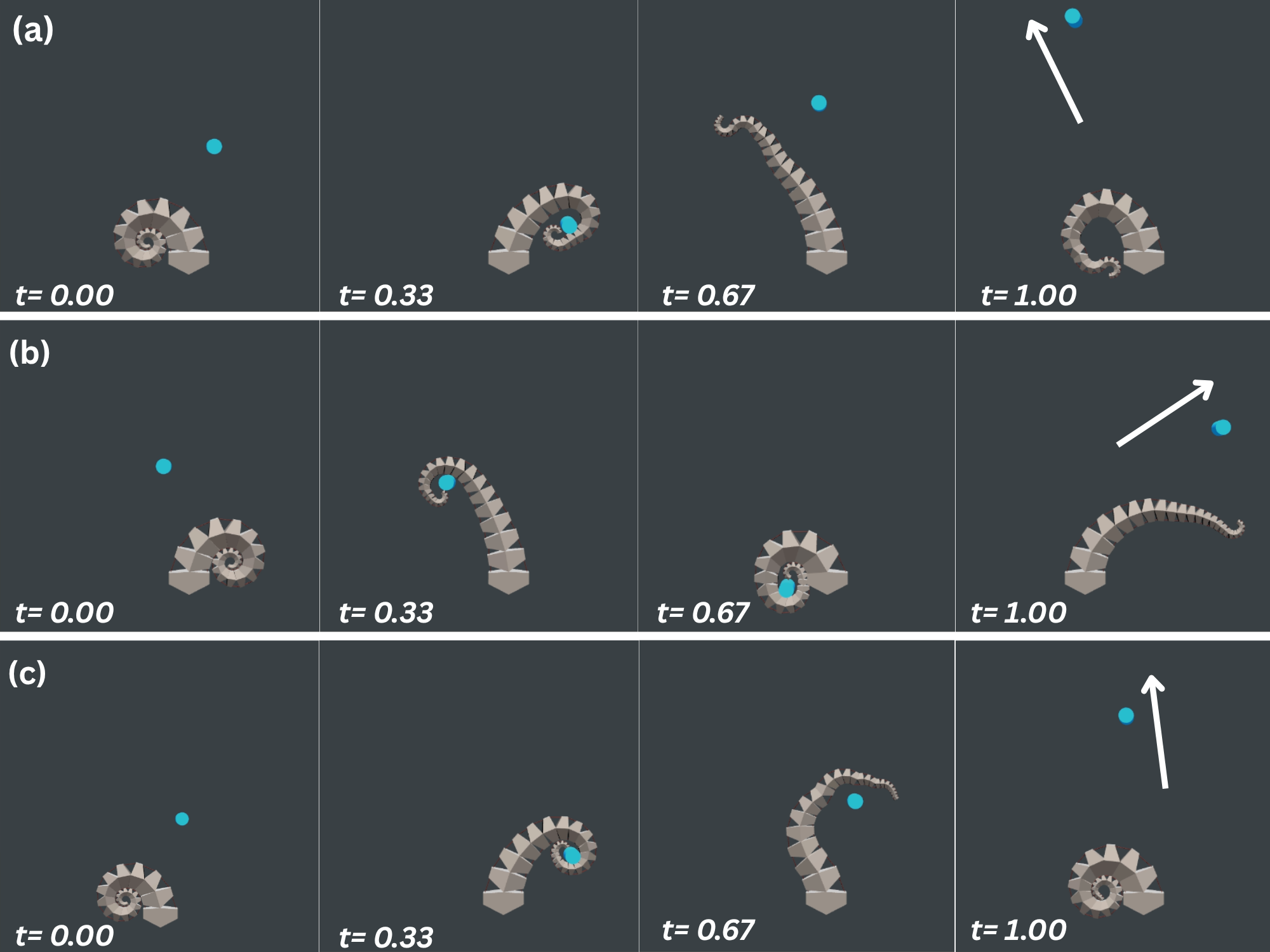}
    \caption{Simulated pick-and-throw trials.}
    \label{fig:simulation_throwing_sequences}
\end{subfigure}

\caption{Simulated task execution sequences. The labeled time $t$ denotes normalized execution time.}
\label{fig:simulation_sequences}
\end{figure}

\newlength{\baselinetablewidth}
\setlength{\baselinetablewidth}{0.825\columnwidth}

\begin{table}[h]
\centering
\caption{Baseline comparisons for whole-body grasping and pick-and-throw.}
\label{tab:throw_baselines}
\scriptsize
\setlength{\tabcolsep}{0.4pt}
\renewcommand{\arraystretch}{1.0}

\textbf{(a) Whole-body grasping}\\[2pt]
\begin{tabular*}{\baselinetablewidth}
{@{\extracolsep{\fill}}l c c c@{}}
\toprule
\textbf{Method} &
\shortstack{\textbf{Grasp success}\\
\textbf{\(n/500\) (\%)}} &
\shortstack{\textbf{Training}\\
\textbf{time (s)}} &
\shortstack{\textbf{Inference}\\
\textbf{time (s)}} \\
\midrule
\textbf{CMA--ES + NN} & \textbf{492/500 (98.4)} & \textbf{41.3} & \textbf{7.57} \\
DE + NN & 345/500 (69.0) & 57.8 & 9.3 \\
SPSA + NN & 250/500 (50.0) & 79.7 & 8.4 \\
TuRBO-1 + NN & 250/500 (50.0) & 79.7 & 8.4 \\
Nelder--Mead + NN & 12/500 (2.4) & 1661.0 & 30.1 \\
\bottomrule
\end{tabular*}

\vspace{6pt}

\textbf{(b) Pick-and-throw Trial 1}\\[2pt]
\begin{tabular}{@{}l c c c c@{}}
\toprule
\textbf{Optimizer + NN}
&
\shortstack{\textbf{Pick-and-throw}\\
\textbf{\(n/100\) (\%)}}
&
\shortstack{\textbf{Direction error}\\
\textbf{\(e_{\mathrm{dir}}\) (\(^{\circ}\))}}
&
\shortstack{\textbf{Training}\\
\textbf{time (s)}}
&
\shortstack{\textbf{Inference}\\
\textbf{time (s)}} \\
\midrule
\textbf{CMA--ES + NN}
& \textbf{98/100 (98.0)} & \textbf{2.4} & \textbf{45.5} & \textbf{8.807} \\
DE + NN
& 74/100 (74.0) & 8.7 & 63.6 & 8.941 \\
SPSA + NN
& 43/100 (43.0) & 14.3 & 87.7 & 8.951 \\
TuRBO-1 + NN
& 47/100 (47.0) & 23.4 & 87.7 & 10.078 \\
Nelder--Mead + NN
& 16/100 (16.0) & 36.0 & 1827.1 & 32.072 \\
\bottomrule
\end{tabular}

\vspace{6pt}

\textbf{(c) Pick-and-throw Trial 2}\\[2pt]
\begin{tabular}{@{}l c c c c@{}}
\toprule
\textbf{Optimizer + NN}
&
\shortstack{\textbf{Pick-and-throw}\\
\textbf{\(n/100\) (\%)}}
&
\shortstack{\textbf{Direction error}\\
\textbf{\(e_{\mathrm{dir}}\) (\(^{\circ}\))}}
&
\shortstack{\textbf{Training}\\
\textbf{time (s)}}
&
\shortstack{\textbf{Inference}\\
\textbf{time (s)}} \\
\midrule
\textbf{CMA--ES + NN}
& \textbf{97/100 (97.0)} & \textbf{2.6} & \textbf{45.5} & \textbf{9.467} \\
DE + NN
& 14/100 (14.0) & 42.2 & 63.6 & 9.699 \\
SPSA + NN
& 14/100 (14.0) & 42.4 & 87.7 & 9.749 \\
TuRBO-1 + NN
& 23/100 (23.0) & 24.8 & 87.7 & 10.559 \\
Nelder--Mead + NN
& 8/100 (8.0) & 42.7 & 1827.1 & 34.941 \\
\bottomrule
\end{tabular}

\vspace{6pt}

\textbf{(d) Pick-and-throw Trial 3}\\[2pt]
\begin{tabular}{@{}l c c c c@{}}
\toprule
\textbf{Optimizer + NN}
&
\shortstack{\textbf{Pick-and-throw}\\
\textbf{\(n/100\) (\%)}}
&
\shortstack{\textbf{Direction error}\\
\textbf{\(e_{\mathrm{dir}}\) (\(^{\circ}\))}}
&
\shortstack{\textbf{Training}\\
\textbf{time (s)}}
&
\shortstack{\textbf{Inference}\\
\textbf{time (s)}} \\
\midrule
\textbf{CMA--ES + NN}
& \textbf{94/100 (94.0)} & \textbf{4.4} & \textbf{45.5} & \textbf{9.065} \\
DE + NN
& 27/100 (27.0) & 28.1 & 63.6 & 9.088 \\
SPSA + NN
& 68/100 (68.0) & 16.5 & 87.7 & 9.068 \\
TuRBO-1 + NN
& 75/100 (75.0) & 12.8 & 87.7 & 10.181 \\
Nelder--Mead + NN
& 23/100 (23.0) & 24.6 & 1827.1 & 32.197 \\
\bottomrule
\end{tabular}
\end{table}

\subsection{Simulation Results}
\label{subsec:ablation}

\begin{figure}[t]
\centering
\includegraphics[width=\columnwidth]{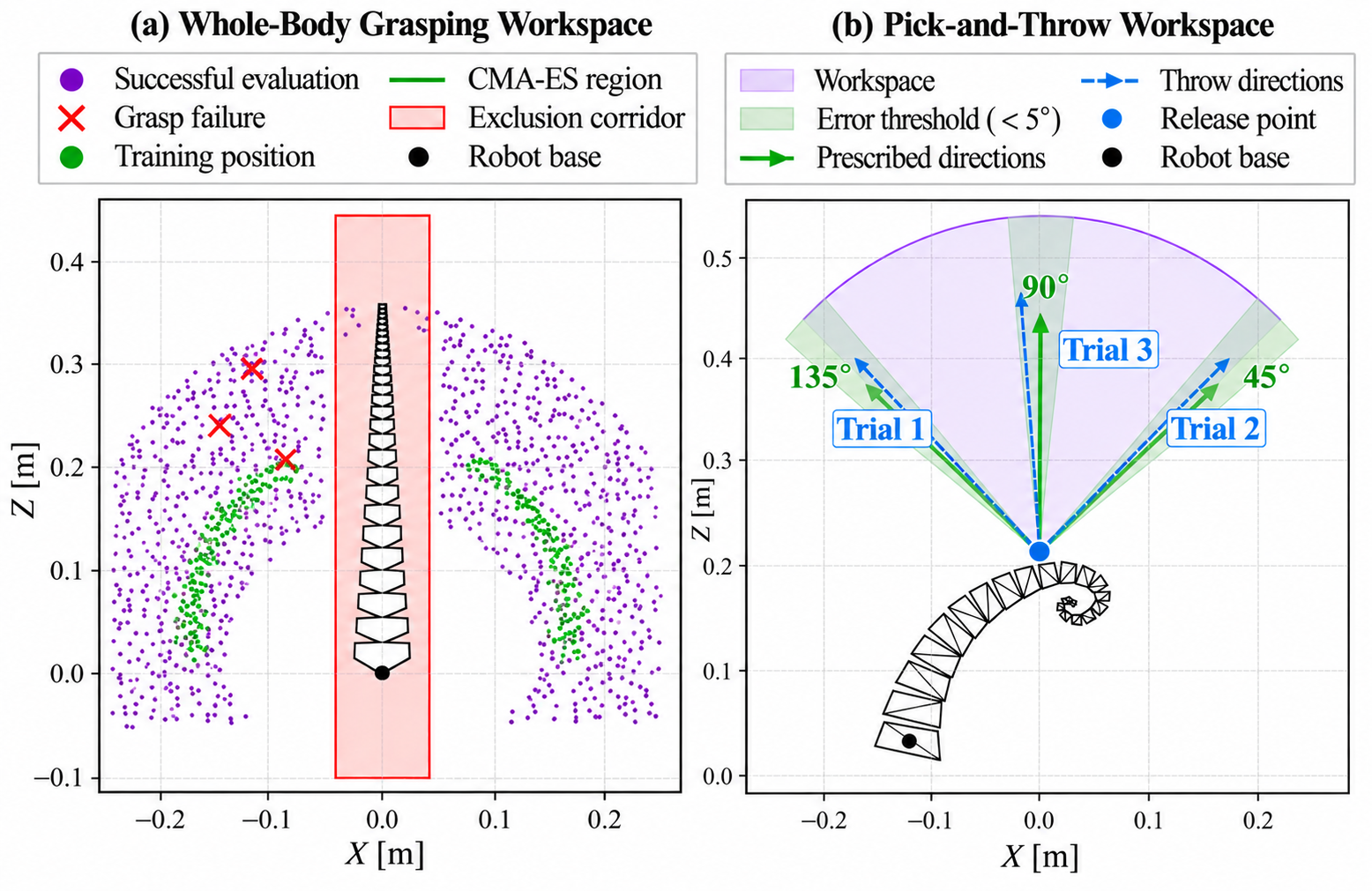}
\caption{Workspace evaluation for
(a) whole-body grasping and (b) directional pick-and-throw.}
\label{fig:workspace}
\end{figure}

Representative simulated whole-body grasping sequences are shown in
Fig.~\ref{fig:simulation_sequences}\subref{fig:simulation_grasping_sequences},
while the workspace-level evaluation is shown in
Fig.~\ref{fig:workspace}(a). As summarized in
Table~\ref{tab:throw_baselines}(a), CMA--ES + NN achieved 492/500
successful randomized grasps ($98.4\%$), substantially higher than the
alternative optimizers. The next-best method, DE + NN, reached only
$69.0\%$, while SPSA + NN and TuRBO-1 + NN both achieved $50.0\%$.
CMA--ES + NN also required the shortest training and inference times.
This indicates that its advantage is not limited to finding more
successful grasps, but also extends to reaching useful solutions more
efficiently across randomized target conditions.

The pick-and-throw results show a similar trend, but more importantly
reveal the consistency of the optimizer across different prescribed
directions. Representative simulated pick-and-throw sequences are shown in
Fig.~\ref{fig:simulation_sequences}\subref{fig:simulation_throwing_sequences},
while the corresponding directional workspace is illustrated in
Fig.~\ref{fig:workspace}(b). As shown in
Table~\ref{tab:throw_baselines}(b)--(d),
CMA--ES + NN maintained success rates of $98\%$, $97\%$, and $94\%$
for Trials~1--3, while keeping the mean direction errors at
$2.4^\circ$, $2.6^\circ$, and $4.4^\circ$, respectively.
Across all three throwing trials, CMA--ES + NN was the only evaluated
method whose mean direction error remained below the $5^\circ$
success threshold, and it also gave the shortest inference time in
each trial. Together with the grasping results, these
comparisons show that CMA--ES provides a more consistent refinement
stage within the evaluated NN-initialized pipelines for both
contact-dependent grasping and directional pick-and-throw.

\begin{table}[b]
\centering
\caption{Ablation comparisons on whole-body grasping.}
\label{tab:grasp_ablation}
\small
\setlength{\tabcolsep}{4pt}
\renewcommand{\arraystretch}{1.08}

\begin{tabular}{@{}l c c@{}}
\toprule
\textbf{Configuration}
& \textbf{Success, \(n/500\) (\%)}
& \shortstack{\textbf{Computation}\\
\textbf{time (s)}} \\
\midrule
\textbf{CMA--ES + NN (proposed)}
& \textbf{492/500 (98.4)}
& \textbf{41.3} \\
\quad without \(A_{\mathrm{wrap}}\)
& 214/500 (42.8)
& 93.0 \\
\quad without \(A_{\mathrm{gap}}\)
& 310/500 (62.0)
& 64.5 \\
\quad without NN
& 393/500 (78.6)
& 73.7 \\
\addlinespace[2pt]
\multicolumn{3}{@{}l@{}}{\footnotesize
Median rollouts: with NN = \textbf{816}; without NN = 1184.} \\
\bottomrule
\end{tabular}
\end{table}

\subsection{Ablation Study}

We further conduct an ablation study to examine the contributions of the
enclosure terms and the learned initialization. Since successful enclosure
is a prerequisite for pick-and-throw, the ablation study is performed on
whole-body grasping, where these effects can be isolated without confounding
grasp failures with subsequent release-direction or speed errors. We evaluate
three ablations of the proposed method:
\begin{itemize}
    \item \textbf{Without $A_{\mathrm{wrap}}$:}
    removes the angular-sweep term from the enclosure objective.

    \item \textbf{Without $A_{\mathrm{gap}}$:}
    removes the body--object proximity term from the enclosure objective.

    \item \textbf{Without NN initialization:}
    replaces the learned task-conditioned initialization with random
    initialization while retaining CMA--ES refinement.
\end{itemize}

As shown in Table~\ref{tab:grasp_ablation}, the two enclosure terms
play complementary roles. Removing $A_{\mathrm{wrap}}$ caused the
largest degradation, reducing grasp success from $98.4\%$ to $42.8\%$,
which indicates that bringing the body close to the object alone is not
sufficient; the robot must also develop sufficient angular coverage to
form a stable whole-body enclosure. Removing $A_{\mathrm{gap}}$ reduced
success to $62.0\%$, showing that angular wrapping without maintaining
close body--object proximity is likewise insufficient for reliable
grasp formation. The learned initialization mainly improves the efficiency and reliability
of refinement rather than replacing optimization itself. Without NN
initialization, CMA--ES still achieved $78.6\%$ success, but required a
higher median rollout count (1184 versus 816) and longer computation
time. This suggests that the neural model provides a task-conditioned
starting point that places CMA--ES closer to a useful solution region,
thereby reducing the search required for condition-specific refinement.


\section{Experimental Validation}
\label{sec:experimental_validation}
\subsection{Experimental Setup}
\label{subsec:experimental_setup}

The SpiRob platform described in Section~\ref{studied-manipulator}
was used for the hardware experiments and mounted above a planar
workspace with two displacement-controlled tendon motors. An overhead
camera recorded the robot--object interaction for offline evaluation
without providing online feedback. The refined tendon commands were
sent to the motors at \(100~\mathrm{Hz}\), with the tendon displacement
rate limited to \(80~\mathrm{mm\,s^{-1}}\). When necessary, the command
profiles were temporally rescaled to satisfy this hardware limit while
preserving their normalized temporal profiles.

The hardware evaluation comprised 50 whole-body grasping executions
and 30 pick-and-throw executions, with 10 repetitions for each of the
three prescribed throwing directions.

\begin{table}[b]
\caption{Hardware validation results. Direction errors and release speeds are averaged over the ten
executions for each prescribed direction.}
\label{tab:hardware_results}
\centering
\footnotesize
\renewcommand{\arraystretch}{1.15}

\textbf{(a) Whole-body grasping}

\vspace{2pt}

\begin{tabular*}{\baselinetablewidth}
{@{\extracolsep{\fill}}l c @{}}

\toprule
\textbf{Task} & \textbf{Success \(n/N\) (\%)} \\
\midrule
Whole-body grasping & 50/50 (100\%) \\
\bottomrule
\end{tabular*}

\vspace{5pt}

\textbf{(b) Pick-and-throw}

\vspace{2pt}

\setlength{\tabcolsep}{2.0pt}
\begin{tabular*}{\baselinetablewidth}
{@{\extracolsep{\fill}}l c c c c @{}}
\toprule
\textbf{Trial} &
\shortstack{\textbf{Target}\\\textbf{direction}} &
\shortstack{\textbf{Direction}\\\textbf{error (\(^{\circ}\))}} &
\shortstack{\textbf{Release speed}\\\textbf{(\(\mathrm{m\,s^{-1}}\))}} &
\shortstack{\textbf{Success}\\\textbf{\(n/N\)} (\%)} \\
\midrule
Trial~1 & \(135^\circ\) & \(4.4^\circ\) & 0.956 & 10/10 (100\%) \\
Trial~2 & \(45^\circ\)  & \(3.8^\circ\) & 0.555 & 10/10 (100\%) \\
Trial~3 & \(90^\circ\)  & \(4.6^\circ\) & 0.827 & 10/10 (100\%) \\
\bottomrule
\end{tabular*}

\end{table}

\subsection{Evaluation Metrics}
\label{subsec:hardware_metrics}

Hardware trial success was determined by manual inspection of the
recorded videos according to the task-specific criteria defined in
Section~\ref{subsec:evaluation_protocol}. For whole-body grasping, a
trial was considered successful if SpiRob captured the target and
retained it at the end of the execution. For pick-and-throw, the video
was inspected to verify the enclosure--acceleration--release sequence
and successful object release.

For quantitative evaluation of pick-and-throw, the camera recordings
were spatially calibrated using the cutting-mat grid. The post-release
object trajectory was tracked to estimate the release velocity. The
resulting release speed and direction error were then computed using
the same definitions and success thresholds as in
Section~\ref{subsec:evaluation_protocol}. A pick-and-throw execution was counted as successful only when the
release sequence was verified from the video and both the prescribed
direction-error and minimum-speed requirements were satisfied.



\subsection{Experimental Results}
\label{subsec:experimental_results}

\begin{figure}[t]
\centering

\begin{subfigure}[t]{\columnwidth}
    \centering
    \includegraphics[width=\columnwidth]{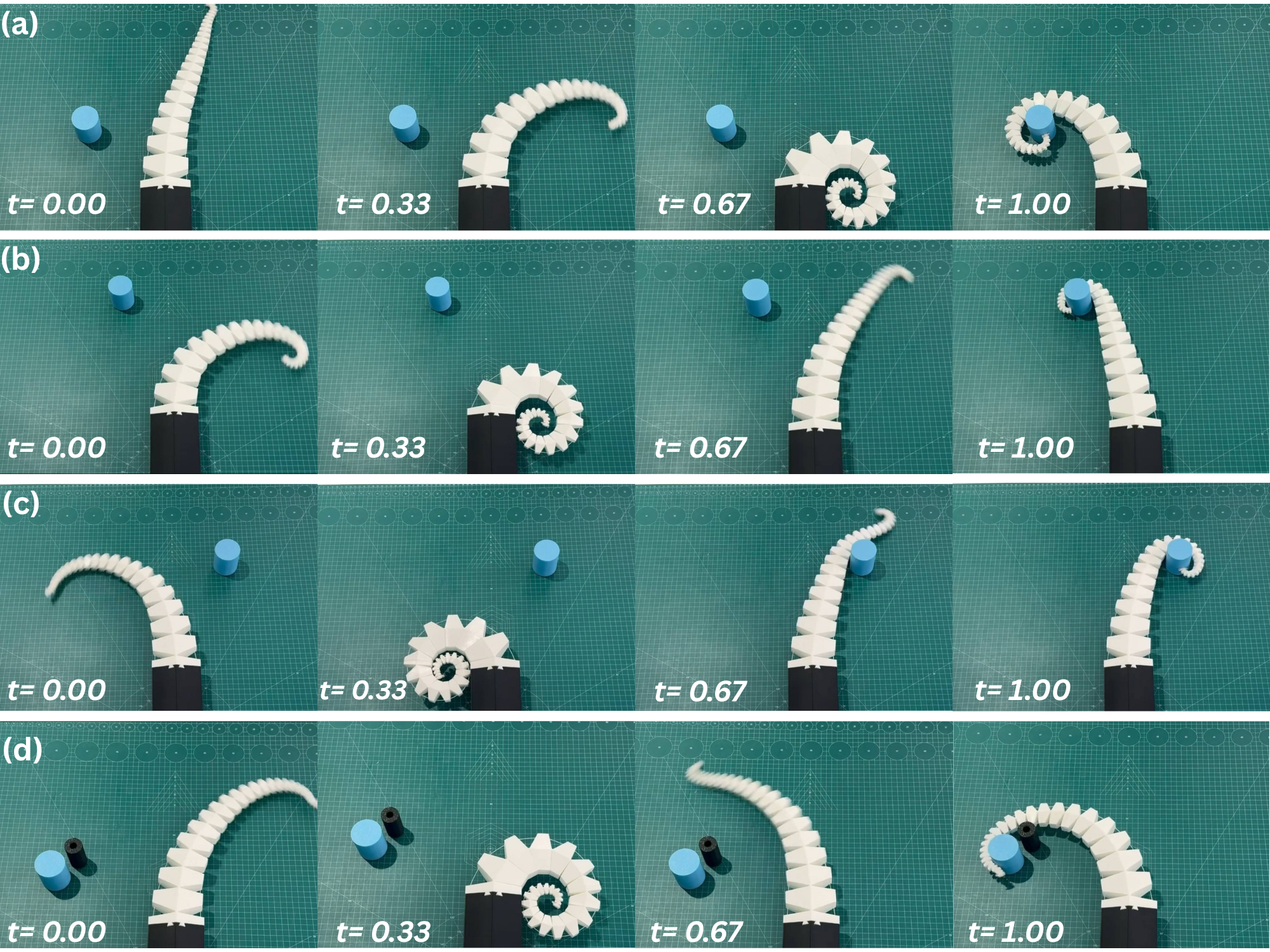}
    \caption{Hardware results for whole-body grasping.}
    \label{fig:hardware_grasp_sequences}
\end{subfigure}

\vspace{2mm}

\begin{subfigure}[t]{\columnwidth}
    \centering
    \includegraphics[width=\columnwidth]{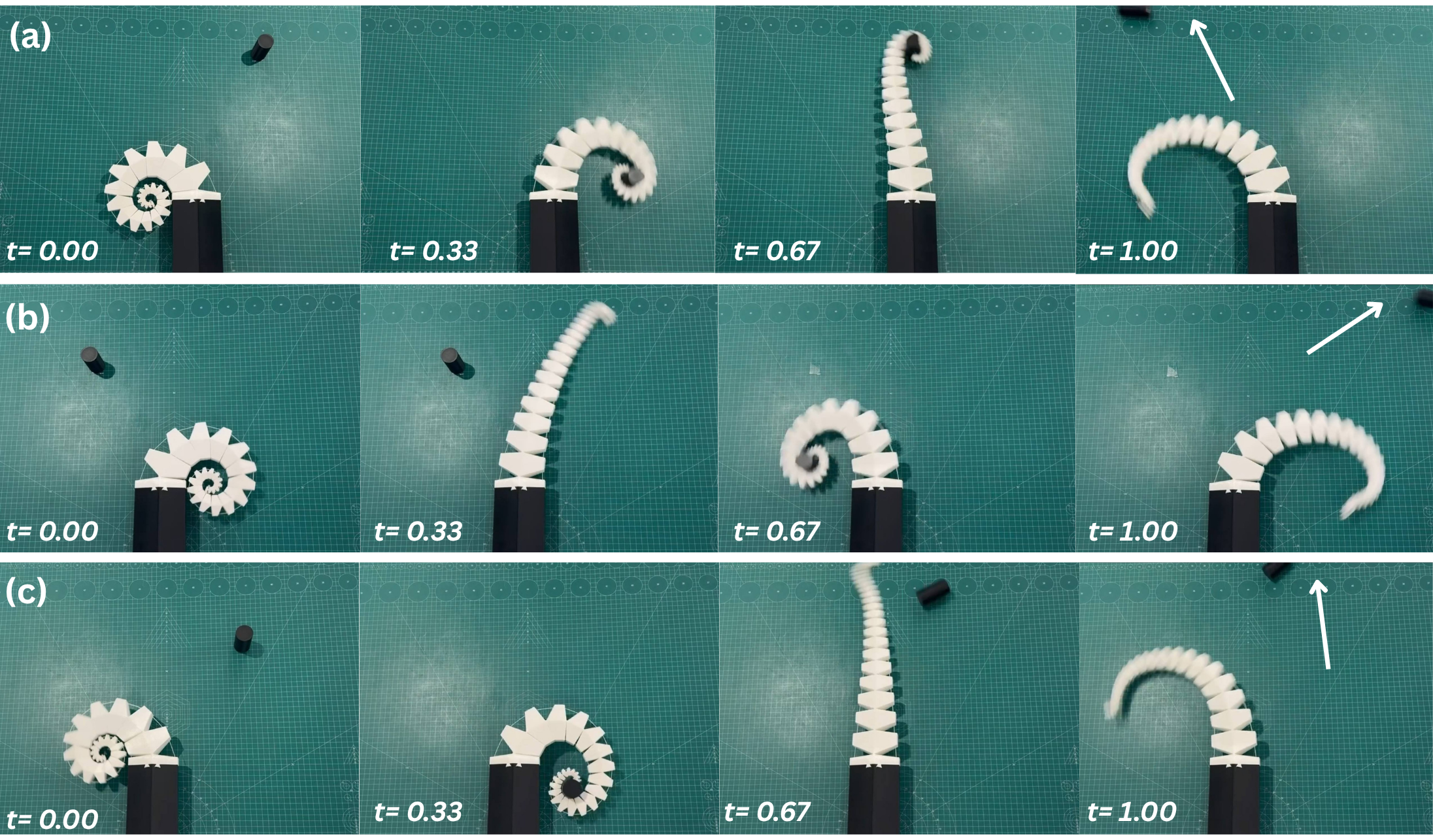}
    \caption{Hardware results for pick-and-throw.}
    \label{fig:hardware_throw}
\end{subfigure}

\caption{Hardware execution sequences for (a) whole-body grasping and
(b) directional pick-and-throw.}
\label{fig:hardware_results}
\end{figure}

\subsubsection{Whole-Body Grasping}
The hardware evaluation comprised 50 grasping executions at different
target positions across the planar workspace. Some trials included an
additional stationary cylindrical object near the target. Representative
sequences from three randomized trials and one trial with the additional
object are shown in Fig.~\ref{fig:hardware_results}(a).



All 50 executions successfully captured and retained the target,
corresponding to a hardware grasping success rate of \(100\%\), as
summarized in Table~\ref{tab:hardware_results}(a).

\subsubsection{Pick-and-Throw}

Representative hardware pick-and-throw sequences for the three
prescribed directions are shown in Fig.~\ref{fig:hardware_results} (b),
with panels (a)--(c) corresponding to Trials~1--3.

Across the 10 executions for each prescribed direction, SpiRob
consistently exhibited the intended enclosure--acceleration--release
sequence, retaining the object during acceleration before release.
All 30 hardware pick-and-throw executions were successful, corresponding
to a \(100\%\) success rate overall and \(10/10\) successful executions
for each direction. The corresponding direction errors and release
speeds are summarized in Table~\ref{tab:hardware_results}(b).

\section{Conclusion}
This work introduced an outcome-based optimize--learn--refine
framework for whole-body grasping and pick-and-throw with
SpiRob. CMA--ES optimizes contact-dependent tendon commands, while a
neural model transfers solutions across conditions for condition-specific
refinement. In simulation, the method achieved a 98.4\% grasping success
rate, reduced the median rollout count from 1184 to 816 in CMA--ES, and achieved
pick-and-throw success rates of 98\%, 97\%, and 94\% across three
throwing directions. On hardware, all 50
whole-body grasping executions were successful, and all 30
pick-and-throw executions, with 10 repetitions per direction,
satisfied the prescribed direction and minimum-speed requirements.
Future work will integrate perception and feedback control to enable
closed-loop three-dimensional grasping and pick-and-throw with SpiRob.

\bibliographystyle{IEEEtran}
\bibliography{references}

\begin{thebibliography}{10}
\providecommand{\url}[1]{#1}
\csname url@samestyle\endcsname
\providecommand{\newblock}{\relax}
\providecommand{\bibinfo}[2]{#2}
\providecommand{\BIBentrySTDinterwordspacing}{\spaceskip=0pt\relax}
\providecommand{\BIBentryALTinterwordstretchfactor}{4}
\providecommand{\BIBentryALTinterwordspacing}{\spaceskip=\fontdimen2\font plus
\BIBentryALTinterwordstretchfactor\fontdimen3\font minus \fontdimen4\font\relax}
\providecommand{\BIBforeignlanguage}[2]{{%
\expandafter\ifx\csname l@#1\endcsname\relax
\typeout{** WARNING: IEEEtran.bst: No hyphenation pattern has been}%
\typeout{** loaded for the language `#1'. Using the pattern for}%
\typeout{** the default language instead.}%
\else
\language=\csname l@#1\endcsname
\fi
#2}}
\providecommand{\BIBdecl}{\relax}
\BIBdecl

\bibitem{mehrkish2021taxonomy}
\BIBentryALTinterwordspacing
A.~Mehrkish and F.~Janabi-Sharifi, ``A comprehensive grasp taxonomy of continuum robots,'' \emph{Robotics and Autonomous Systems}, vol. 145, p. 103860, 2021. [Online]. Available: \url{https://doi.org/10.1016/j.robot.2021.103860}
\BIBentrySTDinterwordspacing

\bibitem{hauser2023morphological}
\BIBentryALTinterwordspacing
H.~Hauser, T.~Nanayakkara, and F.~Forni, ``Leveraging morphological computation for controlling soft robots: Learning from nature to control soft robots,'' \emph{IEEE Control Systems Magazine}, vol.~43, no.~3, pp. 114--129, 2023. [Online]. Available: \url{https://doi.org/10.1109/MCS.2023.3253422}
\BIBentrySTDinterwordspacing

\bibitem{li2011graspconfig}
\BIBentryALTinterwordspacing
J.~Li and J.~Xiao, ``Determining grasping configurations for a spatial continuum manipulator,'' in \emph{Proc. IEEE/RSJ International Conference on Intelligent Robots and Systems}, 2011, pp. 4207--4214. [Online]. Available: \url{https://doi.org/10.1109/IROS.2011.6094663}
\BIBentrySTDinterwordspacing

\bibitem{li2016progressive}
\BIBentryALTinterwordspacing
------, ``Progressive planning of continuum grasping in cluttered space,'' \emph{IEEE Transactions on Robotics}, vol.~32, no.~3, pp. 707--716, 2016. [Online]. Available: \url{https://doi.org/10.1109/TRO.2016.2546308}
\BIBentrySTDinterwordspacing

\bibitem{mehrkish2022synthesis}
\BIBentryALTinterwordspacing
A.~Mehrkish and F.~Janabi-Sharifi, ``Grasp synthesis of continuum robots,'' \emph{Mechanism and Machine Theory}, vol. 168, p. 104575, 2022. [Online]. Available: \url{https://doi.org/10.1016/j.mechmachtheory.2021.104575}
\BIBentrySTDinterwordspacing

\bibitem{graule2022contactimplicit}
\BIBentryALTinterwordspacing
M.~A. Graule, C.~B. Teeple, and R.~J. Wood, ``Contact-implicit trajectory and grasp planning for soft continuum manipulators,'' in \emph{Proc. IEEE/RSJ International Conference on Intelligent Robots and Systems}, 2022, pp. 9401--9408. [Online]. Available: \url{https://doi.org/10.1109/IROS47612.2022.9981044}
\BIBentrySTDinterwordspacing

\bibitem{chu2023fullbody}
\BIBentryALTinterwordspacing
H.~Chu, B.~J. Caasenbrood, M.~Keyvanara, I.~A. Kuling, and H.~Nijmeijer, ``Full-body grasping strategy for planar underactuated soft manipulators using passivity-based control,'' in \emph{Proc. IEEE International Conference on Soft Robotics}, 2023, pp. 1--7. [Online]. Available: \url{https://doi.org/10.1109/RoboSoft55895.2023.10122015}
\BIBentrySTDinterwordspacing

\bibitem{zeng2020tossingbot}
\BIBentryALTinterwordspacing
A.~Zeng, S.~Song, J.~Lee, A.~Rodriguez, and T.~A. Funkhouser, ``{TossingBot}: Learning to throw arbitrary objects with residual physics,'' in \emph{Proceedings of Robotics: Science and Systems}, 2019. [Online]. Available: \url{https://doi.org/10.15607/RSS.2019.XV.004}
\BIBentrySTDinterwordspacing

\bibitem{bianchi2022open}
\BIBentryALTinterwordspacing
D.~Bianchi, M.~G. Antonelli, C.~Laschi, and E.~Falotico, ``Open-loop control of a soft arm in throwing tasks,'' in \emph{Proceedings of the 19th International Conference on Informatics in Control, Automation and Robotics (ICINCO)}, 2022, pp. 138--145. [Online]. Available: \url{https://doi.org/10.5220/0011267100003271}
\BIBentrySTDinterwordspacing

\bibitem{bianchi2024softoss}
\BIBentryALTinterwordspacing
D.~Bianchi, M.~G. Antonelli, C.~Laschi, A.~M. Sabatini, and E.~Falotico, ``{SofToss}: Learning to throw objects with a soft robot,'' \emph{IEEE Robotics \& Automation Magazine}, vol.~31, no.~4, pp. 113--123, 2024. [Online]. Available: \url{https://doi.org/10.1109/MRA.2023.3310865}
\BIBentrySTDinterwordspacing

\bibitem{bianchi2024softsling}
D.~Bianchi, G.~Campinoti, C.~Comitini, C.~Laschi, A.~Rizzo, A.~M. Sabatini, and E.~Falotico, ``{SoftSling}: A soft robotic arm control strategy to throw objects with circular run-ups,'' \emph{IEEE Robotics and Automation Letters}, vol.~9, no.~10, pp. 8250--8257, 2024.

\bibitem{morimoto2022characterization}
\BIBentryALTinterwordspacing
R.~Morimoto, M.~Ikeda, R.~Niiyama, and Y.~Kuniyoshi, ``Characterization of continuum robot arms under reinforcement learning and derived improvements,'' \emph{Frontiers in Robotics and AI}, vol.~9, p. 895388, 2022. [Online]. Available: \url{https://doi.org/10.3389/frobt.2022.895388}
\BIBentrySTDinterwordspacing

\bibitem{dellasantina2023softcontrol}
\BIBentryALTinterwordspacing
C.~D. Santina, C.~Duriez, and D.~Rus, ``Model-based control of soft robots: A survey of the state of the art and open challenges,'' \emph{IEEE Control Systems}, vol.~43, no.~3, pp. 30--65, 2023. [Online]. Available: \url{https://doi.org/10.1109/MCS.2023.3253419}
\BIBentrySTDinterwordspacing

\bibitem{bern2019trajectory}
\BIBentryALTinterwordspacing
J.~M. Bern, P.~Banzet, R.~Poranne, and S.~Coros, ``Trajectory optimization for cable-driven soft robot locomotion,'' in \emph{Proceedings of Robotics: Science and Systems}, 2019. [Online]. Available: \url{https://doi.org/10.15607/RSS.2019.XV.052}
\BIBentrySTDinterwordspacing

\bibitem{menager2025differentiable}
\BIBentryALTinterwordspacing
E.~M{\'e}nager, L.~Montaut, Q.~L. Lidec, and J.~Carpentier, ``Differentiable simulation of soft robots with frictional contacts,'' in \emph{Proc. IEEE International Conference on Soft Robotics}, 2025, pp. 845--852. [Online]. Available: \url{https://doi.org/10.1109/ROBOSOFT63089.2025.11020844}
\BIBentrySTDinterwordspacing

\bibitem{zwane2024dynamic}
\BIBentryALTinterwordspacing
S.~Zwane, D.~G. Cheney, C.~C. Johnson, Y.~Luo, Y.~Bekiroglu, M.~D. Killpack, and M.~P. Deisenroth, ``Learning dynamic tasks on a large-scale soft robot in a handful of trials,'' in \emph{Proc. IEEE/RSJ International Conference on Intelligent Robots and Systems}, 2024, pp. 11\,388--11\,393. [Online]. Available: \url{https://doi.org/10.1109/IROS58592.2024.10802122}
\BIBentrySTDinterwordspacing

\bibitem{shentu2026rigidbody}
\BIBentryALTinterwordspacing
C.~Shentu, N.~Baldassini, T.~Zheng, P.~Rao, and J.~Burgner-Kahrs, ``Do rigid-body simulators dream of soft robots? learning contact-rich manipulation for tendon-driven continuum robots,'' \emph{arXiv preprint arXiv:2606.22397}, 2026. [Online]. Available: \url{https://arxiv.org/abs/2606.22397}
\BIBentrySTDinterwordspacing

\bibitem{wang2024spirobs}
\BIBentryALTinterwordspacing
Z.~Wang, N.~M. Freris, and X.~Wei, ``{SpiRobs}: Logarithmic spiral-shaped robots for versatile grasping across scales,'' \emph{Device}, vol.~3, no.~4, p. 100646, 2025. [Online]. Available: \url{https://doi.org/10.1016/j.device.2024.100646}
\BIBentrySTDinterwordspacing

\bibitem{hansen2016cmaes}
\BIBentryALTinterwordspacing
N.~Hansen, ``The {CMA} evolution strategy: A tutorial,'' \emph{arXiv preprint arXiv:1604.00772}, 2016. [Online]. Available: \url{https://arxiv.org/abs/1604.00772}
\BIBentrySTDinterwordspacing

\bibitem{hansen2001cmaes}
N.~Hansen and A.~Ostermeier, ``Completely derandomized self-adaptation in evolution strategies,'' \emph{Evolutionary Computation}, vol.~9, no.~2, pp. 159--195, 2001.

\bibitem{storn1997differential}
R.~Storn and K.~Price, ``Differential evolution---a simple and efficient heuristic for global optimization over continuous spaces,'' \emph{Journal of Global Optimization}, vol.~11, pp. 341--359, 1997.

\bibitem{spall1992spsa}
J.~C. Spall, ``Multivariate stochastic approximation using a simultaneous perturbation gradient approximation,'' \emph{IEEE Transactions on Automatic Control}, vol.~37, no.~3, pp. 332--341, 1992.

\bibitem{eriksson2019turbo}
D.~Eriksson, M.~Pearce, J.~R. Gardner, R.~D. Turner, and M.~Poloczek, ``Scalable global optimization via local {Bayesian} optimization,'' in \emph{Advances in Neural Information Processing Systems}, vol.~32, 2019, pp. 5496--5507.

\bibitem{nelder1965simplex}
J.~A. Nelder and R.~Mead, ``A simplex method for function minimization,'' \emph{The Computer Journal}, vol.~7, no.~4, pp. 308--313, 1965.

\end{thebibliography}
\end{document}